\documentclass{article} 
\usepackage{iclr2027_conference,times}

\usepackage{amsmath,amsfonts,bm}

\def\eqref#1{equation~\ref{#1}}

\def\1{\bm{1}}

\DeclareMathAlphabet{\mathsfit}{\encodingdefault}{\sfdefault}{m}{sl}
\SetMathAlphabet{\mathsfit}{bold}{\encodingdefault}{\sfdefault}{bx}{n}

\usepackage{hyperref}
\usepackage{url}
\usepackage{graphicx}
\usepackage[caption=false,font=footnotesize]{subfig}
\usepackage{booktabs}
\usepackage{tcolorbox}
\usepackage{array}
\usepackage{longtable}
\usepackage{wrapfig}        
\usepackage{xcolor,colortbl}
\usepackage{color}
\tcbuselibrary{breakable}

\title{ReflectPO: Reflection-Guided Policy Optimization for Long-Horizon LLM Agents}

\author{Rumei Li$^1$ \ , Xunliang Cai$^1$\\
$^1$Meituan, China\\
\texttt{lirumei@meituan.com}\\
\\
}
\iclrfinalcopy 
\begin{document}

\maketitle
\fancyhead{}

\begin{abstract}

With external tools, large language models (LLMs) can tackle complex tasks through sustained environmental interaction. LLM-based agents are therefore increasingly used to tackle complex long-horizon tasks in real-world environments. However, these tasks involve sequential, interdependent decisions and changes in environmental states. Early execution errors affect subsequent interaction histories and environmental states, and their effects accumulate along the decision chain. To address this challenge, we propose Reflection-Guided Policy Optimization (ReflectPO), a reinforcement learning framework for autonomous error recovery in long-horizon tasks. ReflectPO comprises two core components. The first is Reflection-Guided Sampling. We use branch differences and repeated visits in interaction trajectories to construct hints for error recovery. The second is Tree-Structured Credit Assignment. We organize sampled trajectories into trees and assign action-level credit using parent--child reward differences and recovery gains. Experiments over two long-horizon tool-use benchmarks show that ReflectPO outperforms multiple existing baselines in most domains. It also achieves the best overall performance on average. Further analyses show that ReflectPO effectively improves agents' ability to recognize their own deviations, recover from erroneous states, and continue making progress toward task completion.

\end{abstract}

\section{Introduction}

Large language models (LLMs) have made steady progress in tasks such as reasoning, instruction following, task planning, and tool use~\citep{deepseek-r1,autoif,plan-and-act,toolstar,webthinker,swe-rl}. This progress has also supported the development of LLM-based agents. LLM-based agents use tools to interact with external environments and turn model decisions into actions for complex real-world tasks~\citep{webagent-r1,agentscaler,LOOP,retool,t-bench,mcp-atlas,mcp-universe,tool-decathlon}. Such tasks involve long-horizon interactions in which decisions shape subsequent states and interaction histories~\citep{longhorizonsurvey}. These dependencies complicate state tracking and decision-making.

%
To improve agents' ability to handle these long-horizon dependencies, reinforcement learning methods optimize agent policies through multi-turn agent--environment interactions, with designs targeting trajectory exploration, credit assignment, and error recovery~\citep{arpo,et-agent,ToolOrchestra,beacon}. Despite these performance gains, execution errors can still alter subsequent states and interaction histories, causing deviations to accumulate and tasks to fail~\citep{rupa,fama}. Detecting and recovering these errors remains difficult, making autonomous recovery a key challenge for reliable long-horizon execution.

To support recovery from errors during long-horizon execution, existing methods provide inference-time assistance or train agents to recover from failed attempts. Inference-time recovery methods use language feedback, memory, or external verification to assess ongoing execution~\citep{reflexion,critic}. However, these methods often require additional external assistance at test time. Training-based recovery methods derive learning signals from failed trajectories, step-level feedback, or diagnostic information~\citep{agent-r,steca,fission-grpo}. However, their reliance on externally provided error locations or recovery guidance limits supervision for deciding when to initiate recovery independently. Meanwhile, general-purpose reinforcement learning primarily optimizes task outcomes, but these outcome-based signals provide indirect guidance for error detection and recovery~\citep{arpo,aepo,et-agent}. Prior studies also show that better task-solving performance does not automatically improve self-verification~\citep{self-verify,liu2026llm}. Overall, these approaches provide limited support for learning autonomous error recovery. Therefore, how to enable agents to identify their own errors and recover from deviations remains an open challenge.

In this paper, we propose Reflection-Guided Policy Optimization (ReflectPO), a reinforcement learning method for error recovery in long-horizon agent tasks. Specifically, ReflectPO introduces two recovery trajectory sampling strategies: contrast-based sampling and loop-based sampling. The former organizes trajectories for the same task into a shared prefix tree, and constructs hints from differences between execution paths. The latter constructs a local interaction graph for each trajectory, and derives hints from loops within that trajectory. Guided by these hints, the agent resumes the task from recovery anchors and generates recovery trajectories.

To derive precise policy update signals from sampled trajectories, we design a tree-structured credit assignment method that accounts for both the effects of local decisions and the gains from recovery. These training signals reinforce effective task execution and error recovery, improving agents' ability to complete long-horizon tasks.

We evaluate ReflectPO on two tool-use benchmarks. ReflectPO consistently improves average performance on both benchmarks. Further analyses suggest that trained agents can autonomously identify execution errors and recover from deviations to some extent, demonstrating its effectiveness in long-horizon agent tasks.

In summary, our main contributions are as follows:
\begin{itemize}
    \item We propose ReflectPO, which uses directed graphs to extract local differences and repeated visitation patterns from natural reasoning trajectories. Two sampling strategies use these patterns to guide agents in generating recovery trajectories.
    \item We propose tree-structured credit assignment based on the structure of sampled trajectories. It more accurately captures the effects of local decisions on subsequent outcomes and the actual gains from recovery.
    \item We conduct systematic experiments on two challenging long-horizon tasks. ReflectPO improves average performance over conventional methods. Further analyses show that trained agents exhibit clear error recovery behaviors.
\end{itemize}

\section{Related Work}
\label{sec:related_work}

\subsection{Failures and Recovery in Long-Horizon Agent Execution}

During long-horizon tasks, agents can make errors in planning, memory, state tracking, and tool interaction. These errors can accumulate across interdependent decisions. AgentDebug, TRAJECT-Bench, LUMINA, and HORIZON diagnose these failures and performance degradation~\citep{agentdebug,trajectbench,lumina,horizon}. Longer action sequences complicate exploration and credit assignment and amplify short-term decision errors~\citep{horizonlength,whyreasoningfailstoplan}, while ReflecTool-Bench reveals weaker detection and recovery of agents' own errors than recognition of external ones~\citep{liu2026llm}. Zombie Agents and AgenTracer support recovery through loop monitoring and failure attribution~\citep{zombieagent,agenttracer}, but do not directly train autonomous deviation detection and correction. We use branch outcome differences and repeated visitation patterns in natural trajectories to train reflection and recovery.


\subsection{Agentic Reinforcement Learning for Long-Horizon Tasks}

Agentic reinforcement learning optimizes long-horizon exploration, credit assignment, and recovery using interaction outcomes~\citep{retool,gigpo,webagent-r1,ragen}. SPEAR, ARPO, and Tree-GRPO improve sampling and step-level advantage estimation through progressive exploration, branch selection, or shared trajectory structures~\citep{spear,arpo,appo,treegrpo,salt,g2po,hgpo}. RLVMR, Fission-GRPO, and P-BRIDGE learn reflection and error recovery from verifiable rewards, failures, diagnostic feedback, or teacher corrections~\citep{rlvmr,fission-grpo,regrpo,p-bridge}. These approaches give limited attention to discovering latent deviations in natural rollouts and learning when to reflect. We use branch differences and repeated visitation patterns to identify key states and construct recovery trajectories, training the policy to reflect and recover autonomously at inference time.

\section{Preliminary}
\label{sec:preliminary}

\subsection{Problem Formulation}
\label{ssec:problem_formulation}

Following prior work\citep{agentworld,envscaler,envfactory}, we model long-horizon tool interaction as a task-conditioned, partially observable process. Each task $x\sim\mathcal{D}$ has an initial state $s_0^x$, an environment $\mathcal{E}_x$, and a context $c_x$ specifying instructions, constraints, and tool interfaces. The agent policy $\pi_\theta$ operates over $\mathcal{A}=\mathcal{A}_{\mathrm{tool}}\cup\mathcal{A}_{\mathrm{resp}}$, comprising structured tool calls and natural language responses. 



The agent samples $a_t\sim\pi_\theta(\cdot\mid h_t)$ from the history $h_t=(c_x,a_0,o_0,\ldots,a_{t-1},o_{t-1})$, with $h_0=c_x$. $o_t$ represents tool execution results or user messages. The environment executes $a_t$, updating state and returning an observation according to $(s_{t+1},o_t)\sim\mathcal{E}_x(\cdot\mid s_t,a_t)$. 
Interaction continues until a final response or the step limit, yielding a trajectory:
\begin{equation}
    \tau=(a_0,o_0,a_1,o_1,\ldots,a_T,o_T),
\end{equation}
Here, $a_T$ is the final response or termination action, and $o_T$ is the terminal observation. Earlier actions shape subsequent decisions through environment states and interaction history, allowing errors to accumulate or be recovered by later actions. We examine these error and recovery patterns in natural agent trajectories.


\subsection{Preliminary Experiment}
\label{ssec:preliminary_experiment}

To characterize errors and recovery during normal execution, we analyze long-horizon trajectories generated by Qwen3-8B on $\tau^2$-bench and BFCL-v3 Multi-turn~\citep{tau2-bench,bfcl}. The model follows its standard reasoning and tool-call procedure to generate trajectories.

We use \texttt{GPT-5.6 Terra} as the judge model~\footnote{https://openai.com/index/gpt-5-6/. }. 
The judge identifies all steps with explicit errors and determines whether the agent successfully recovers each error in later steps. We report four metrics: error trajectory rate, the fraction of trajectories containing errors; error step rate, the fraction of visible steps with errors; step recovery rate, the fraction of erroneous steps later recovered; and complete recovery rate, the fraction of error-containing trajectories with all errors corrected. We repeat each evaluation three times and report the average results~\footnote{Judge prompts are provided in Appendix~\ref{app:preliminary_judge_prompt}.}.

\begin{table}[t]
    \centering
    \renewcommand{\tablename}{Table}
    \setlength{\belowcaptionskip}{6pt}
    \def\baselinestretch{1}
    \fontsize{8.5pt}{11pt}\selectfont
    \setlength{\tabcolsep}{2pt}
    \renewcommand{\arraystretch}{1.15}
    \caption{Errors and autonomous recovery in natural agent trajectories.}
    \label{tab:preliminary_error_recovery}
    \begin{tabular*}{\linewidth}{@{}l@{\extracolsep{\fill}}cccc@{}}
        \toprule
        Dataset & Error Trajectory Rate & Error Step Rate & Step Recovery Rate & Complete Recovery Rate \\
        \midrule
        Overall & 83.68 & 19.49 & 18.74 & 14.19 \\
        BFCL-v3 & 79.38 & 16.27 & 19.18 & 17.95 \\
        $\tau^2$-Bench & 96.04 & 28.74 & 17.48 & 3.37 \\
        \bottomrule
    \end{tabular*}
\end{table}


The result is shown in Table~\ref{tab:preliminary_error_recovery}. We can draw two conclusions. \textbf{(1) Errors are widespread.} Error trajectory rates reach 79.38 on BFCL-v3 and 96.04 on $\tau^2$-bench, indicating that errors occur across different types of multi-turn agent tasks. \textbf{(2) Autonomous error recovery remains limited.} The step recovery rate is 18.74, while the complete recovery rate is 14.19. These results suggest that the base policy does not consistently detect errors and recover from them. Our method substantially alleviates this limitation. It incorporates reflection-guided recovery trajectories into policy optimization to help agents learn to recover from errors and resume task progress.

\section{Method}
\label{sec:method}

\textbf{Method Overview:} This section presents Reflection-Guided Policy Optimization (ReflectPO), a reinforcement learning method for error recovery in long-horizon tasks. ReflectPO identifies execution points in original interaction trajectories that require further review. It then generates recovery trajectories from the corresponding environment states. The method has two core components:
\begin{itemize}
    \item \textbf{Reflection-Guided Sampling (\S\ref{ssec:reflexion_sampling}).} We design contrast-based and loop-based sampling to identify local trajectory differences and repeated visits. A guide model converts these signals into reflection hints that guide the agent to generate recovery trajectories.
    \item \textbf{Tree-Structured Credit Assignment (\S\ref{ssec:tree_credit}).} We build a full prefix tree from original trajectories and compute step-level action advantages from terminal outcomes across its branches. Each recovery trajectory is paired with its target original trajectory. Their reward difference determines the advantage for recovery actions.
\end{itemize}

\subsection{Interaction Node Representation}
\label{ssec:interaction_node}


Before introducing these components, we define the interaction node representation. Following the interaction process in Section~\ref{ssec:problem_formulation}, we represent the agent action at step $t$ and its corresponding environment observation as $\xi_t=\langle a_t,o_t\rangle$. 
Given a window length $w>0$, the interaction node at step $t$ contains the most recent $w$ action--observation pairs, shown as
$v_t^{(w)}=\left(\xi_{\max(0,t-w+1)},\ldots,\xi_t\right)$.
When $w=t+1$, the node retains the entire interaction from the start of the task, denoted as $v_t^{(\mathrm{full})}$.
We normalize actions and observations to construct a signature for each action--observation pair. A structured tool action is represented by its tool name and normalized arguments. We remove nondeterministic differences in tool observations, such as timestamps, random identifiers, and field order, while preserving execution results. Two nodes match if they belong to the same task, contain the same number of action--observation pairs, and have identical corresponding signatures.

For a node ending in $\xi_t$, we save an environment snapshot after the last action. The snapshot records the full internal state $s_{t+1}$, including user-simulator or dialogue-environment information when applicable. Reflection-guided sampling restores this snapshot to resume execution.

\subsection{Reflection-Guided Sampling}
\label{ssec:reflexion_sampling}

\begin{figure}[t]
    \centering
    \includegraphics[width=\linewidth]{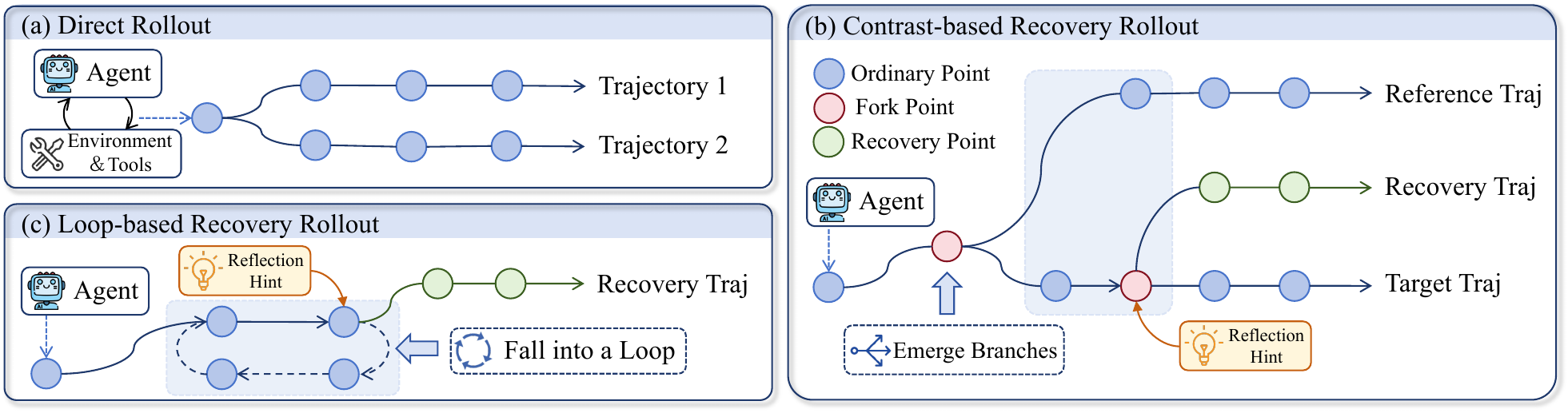}
    \caption{Recovery trajectory sampling in ReflectPO. Contrast-based sampling compares target and reference trajectories after a shared branch point. Loop-based sampling identifies repeated visits within a trajectory. Both insert reflection hints at recovery anchors and generate new rollouts.}
    \label{fig:recovery_sampling}
\end{figure}

As shown in Section~\ref{ssec:preliminary_experiment}, the base policy makes frequent errors during long-horizon interactions and has limited autonomous recovery. To generate trajectories with reflection and recovery behaviors, we design contrast-based sampling and loop-based sampling strategies. We use a stronger LLM $\pi_{\mathrm{guide}}$ as a guide model. Given task information and selected interaction segments, $\pi_{\mathrm{guide}}$ generates reflection hints to guide the behavior policy in replanning subsequent execution. Figure~\ref{fig:recovery_sampling} illustrates both sampling strategies.

\paragraph{Contrast-based sampling.}
Let $\mathcal{T}_x^{0}=\{\tau_i^{0}\}_{i=1}^{K}$ be the original trajectory set for task $x$. Let $T_i$ be the terminal step of trajectory $\tau_i^0$ and $r_i^0$ its terminal reward. We represent $\tau_i^0$ as a sequence of full prefix nodes $\{v_{i,t}^{(\mathrm{full})}\}_{t=0}^{T_i}$. The node sequences for the same task form a prefix tree $\mathcal{P}_x^0$. Trajectories with identical full prefixes share tree nodes. Nodes with multiple distinct successors serve as candidate branch points. For each candidate node $v$, we extract a local segment of at most $H_c$ steps after the node from each member trajectory. Identical segments are grouped together. Let $\mathcal{G}(v)$ denote the resulting groups and $\bar r_g$ the mean terminal reward of group $g$. We define the outcome difference at node $v$ as:
\begin{equation}
    D_{\mathrm{contrast}}(v)
    =\operatorname{Var}\left(\{\bar r_g:g\in\mathcal{G}(v)\}\right).
    \label{eq:contrast_score}
\end{equation}
A larger $D_{\mathrm{contrast}}(v)$ indicates greater outcome differences across local behaviors following the node. We prioritize nodes with larger scores when constructing contrast-based recovery trajectories.

For each selected branch point, we choose the member trajectory with the lowest terminal reward as the target trajectory to be guided toward recovery. We select another member trajectory with a different local segment and a terminal reward no lower than the target's as the reference trajectory for comparison. 
Given the task context, shared prefix, and the two local segments, the guide model generates a reflection hint that identifies target decisions to review and provides guidance for subsequent planning. We search backward from the end of the target segment for the latest node with a valid environment snapshot and use it as the recovery anchor.

\paragraph{Loop-based sampling.}
We use a fixed window length $w_{\ell}>0$ to define local interaction nodes for loop detection. Each original trajectory $\tau_i^0$ is represented as a sequence of finite-window nodes $\{v_{i,t}^{(w_{\ell})}\}$. We connect adjacent nodes in visitation order to form a directed graph. When a node is visited again, the path between its two visits forms a candidate loop segment. Graph construction and loop detection are performed independently for each original trajectory.

For the first repeated node, we extract the complete interaction segment from the start of its first visit to the end of its second visit. The guide model $\pi_{\mathrm{guide}}$ generates a reflection hint from the task information, the context before the loop, and the loop segment. The hint summarizes the repeated behavior and suggests an actionable alternative. The end of the second visit contains the environment feedback and internal state after loop execution. We use this position as the loop recovery anchor.

\paragraph{Recovery trajectory construction.}
We restore the target trajectory's environment snapshot at the recovery anchor, which records the state after the anchor action, and insert the reflection hint after the corresponding observation. Conditioned on the task context, interaction prefix, and hint, the behavior policy $\pi_{\theta_{\mathrm{old}}}$ resumes interaction, appending each new action and observation until the task terminates. The task reward mechanism assigns the resulting recovery trajectory a terminal reward. Appendix~\ref{app:prompt_details} provides the guide prompts. We next define step-level advantages from the original trajectory tree and paired recovery outcomes.

\subsection{Tree-Structured Credit Assignment}
\label{ssec:tree_credit}

\begin{figure}[t]
    \centering
    \includegraphics[width=\linewidth]{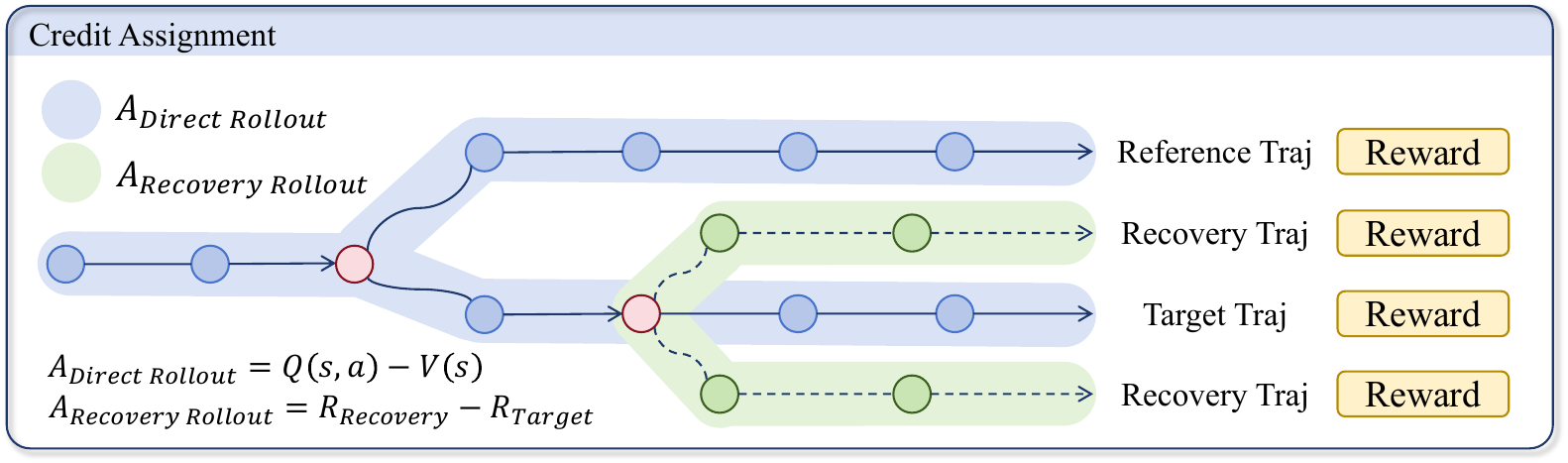}
    \caption{Credit assignment in ReflectPO. Original actions receive tree-based advantages from empirical outcome differences between parent and child nodes. Recovery actions receive paired advantages from terminal reward differences between recovery and target original trajectories.}
    \label{fig:credit_assignment}
\end{figure}

To estimate action advantages accurately in three types of trajectories, we design a tree-structured credit assignment method. Let $\mathfrak{B}$ denote the current global training batch, with original trajectories $\mathcal{T}_x^0$ and recovery trajectories $\mathcal{T}_x^{\mathrm{rec}}$ for each task $x$. We assign tree-based advantages to original actions and paired reward-difference advantages to recovery actions. Figure~\ref{fig:credit_assignment} illustrates this process.

\paragraph{Expanded batch construction.}
We combine all original and recovery trajectories into an expanded training batch, then pad it once for distributed execution. Padding rows have zero training weight and are excluded from credit assignment and loss averaging. Appendix~\ref{app:batch_alignment} gives the alignment and distributed normalization details.

\paragraph{Direct Trajectory Attribution.}
Inspired by the tree-based Monte Carlo estimator in RTMC\citep{RTMC}, we estimate advantages on the full prefix tree $\mathcal{P}_x^0$ constructed from original trajectories $\mathcal{T}_x^0$. Let $\mathcal{I}(v)$ index the trajectories passing through node $v$. A branch node has at least two distinct children. For an action $a$ taken at a branch node $v$, let $u$ be the child reached by that transition. Its raw advantage is the difference in mean terminal reward between trajectories passing through $u$ and those passing through $v$:
\begin{equation}
    A^{\mathrm{raw}}(v,a)
    =\frac{1}{|\mathcal{I}(u)|}\sum_{i\in\mathcal{I}(u)}r_i^0
    -\frac{1}{|\mathcal{I}(v)|}\sum_{i\in\mathcal{I}(v)}r_i^0.
    \label{eq:direct_edge_advantage}
\end{equation}

\paragraph{Batch-Level Normalization.}
Let $\sigma_{\mathfrak{B}}$ denote the standard deviation of raw advantages from branch-node visits by original trajectories. It is computed over all such visits in the current global batch. Each visit contributes separately, including repeated visits across trajectories.
Let $v_{i,t}^{-}$ denote the full prefix node before action $a_{i,t}$. We scale and clip its raw advantage as
\begin{equation}
    \widehat A_{i,t}^{\mathrm{dir}}=
    \begin{cases}
        \operatorname{clip}\!\left(\frac{A^{\mathrm{raw}}(v_{i,t}^{-},a_{i,t})}{\sigma_{\mathfrak{B}}},-1,1\right),
        & \text{if }v_{i,t}^{-}\text{ branches},\\
        \widehat A_{i,t-1}^{\mathrm{dir}},
        & \text{otherwise, if }t>0,\\
        0, & \text{otherwise, if }t=0.
    \end{cases}
    \label{eq:direct_advantage_assignment}
\end{equation}
Actions along a single path inherit the most recent branch advantage, with zero advantage before the first branch. Each new branch updates the inherited value.

\paragraph{Paired Recovery Attribution.}
For the $m$-th recovery trajectory in $\mathcal{T}_x^{\mathrm{rec}}$, let $\widetilde r_m$ denote its terminal reward and $q(m)$ the index of its target original trajectory. Its recovery advantage is
\begin{equation}
    \widehat A_m^{\mathrm{rec}}
    =\operatorname{clip}\!\left(\frac{\widetilde r_m-r_{q(m)}^0}{\sigma_{\mathfrak{B}}},-1,1\right).
    \label{eq:recovery_advantage}
\end{equation}
All agent actions newly generated after the recovery anchor share $\widehat A_m^{\mathrm{rec}}$.

\paragraph{Policy Optimization.}
Let $n$ index batch trajectories, $j$ index tokens, $y_{n,j}$ the corresponding token, and $\widehat A_{n,j}$ denote the token advantage. We optimize $\pi_\theta$ using trajectories generated by $\pi_{\theta_{\mathrm{old}}}$. Following prior work, we remove hints from recovery sequences during updates and compute gradients using contexts without hints~\citep{ruscarl,agent0}. The token probability ratio is

\begin{equation}
    \rho_{n,j}(\theta)
    =\frac{\pi_\theta(y_{n,j}\mid y_{n,<j})}
    {\pi_{\theta_{\mathrm{old}}}(y_{n,j}\mid y_{n,<j})}.
\end{equation}


To limit the influence of trajectory and action lengths, we average equally over trajectories, steps, and tokens. Let $\mathcal{B}_{\mathrm{valid}}$ index valid real trajectories with at least one trainable step in the global batch. For trajectory $n$, let $S_n$ count its trainable assistant steps and $\mathcal{J}_{n,s}$ contain valid model-token positions in step $s$. Environment observations, guide hints, fixed shared prefixes of recovery trajectories, and padding tokens are excluded. Valid steps with zero advantage still count toward $S_n$.

Let $\epsilon>0$ be the clipping threshold for the policy ratio. We use these token-level advantages in the objective adopted by Reinforce++\citep{hu2025reinforcestabilizingcriticfreepolicy}. The loss for a single token is
\begin{equation}
\begin{split}
    \ell_{n,j}(\theta)
    =-\min\Big(&
    \rho_{n,j}(\theta)\widehat A_{n,j},\\
    &\operatorname{clip}(\rho_{n,j}(\theta),1-\epsilon,1+\epsilon)
    \widehat A_{n,j}
    \Big).
\end{split}
    \label{eq:arpo_token_loss}
\end{equation}

The final optimization objective is
\begin{equation}
    \mathcal{L}_{\mathrm{ReflectPO}}(\theta)
    =\frac{1}{|\mathcal{B}_{\mathrm{valid}}|}
    \sum_{n\in\mathcal{B}_{\mathrm{valid}}}
    \frac{1}{S_n}\sum_{s=1}^{S_n}
    \frac{1}{|\mathcal{J}_{n,s}|}\sum_{j\in\mathcal{J}_{n,s}}
    \ell_{n,j}(\theta) + \beta_{\mathrm{KL}}\mathcal{L}_{\mathrm{KL}}(\theta),
    \label{eq:arpo_objective}
\end{equation}
where $\mathcal{L}_{\mathrm{KL}}$ denotes KL regularization. This objective combines tree-based advantages for original actions with paired recovery advantages under a shared batch-level scale.



\section{Experiments}
\label{sec:experiments}

\subsection{experimental setup}
\label{sec:experimental_setup}

\noindent\textbf{Training Details.}
We train Qwen3 models with ReflectPO~\citep{qwen3}, using RL data from EnvScaler~\citep{envscaler}. 
Following the evaluation setup of $\tau^2$-Bench, we configure both conversation and non-conversation modes for training.
Appendix~\ref{app:details} provides further details.

\noindent\textbf{Benchmarks.}
We evaluate multi-turn tool use on BFCL-v3 Multi-Turn and $\tau^2$-Bench~\citep{bfcl,tau2-bench}. BFCL-v3 Multi-Turn assesses multi-turn dialogue, function calling, and context handling in stateful API environments. $\tau^2$-Bench evaluates agents' ability to follow domain rules, invoke tools, and complete tasks during user conversations.

\noindent\textbf{Baselines.}
We compare direct inference with Qwen3-4B and Qwen3-8B against results obtained using PPO, GRPO, Reinforce++, GiGPO, CRITIC, and Fission-GRPO. Detailed descriptions of these baselines are provided in Appendix~\ref{app:baselines}.


\subsection{Main Results}
\label{ssec:main_results}

Table~\ref{tab:main_results} presents the experimental results. Overall, ReflectPO performs strongly across multiple task domains and achieves the best average performance. We draw three main conclusions.

\begin{wrapfigure}[14]{r}{0.50\textwidth}
    \centering
    \includegraphics[width=\linewidth]{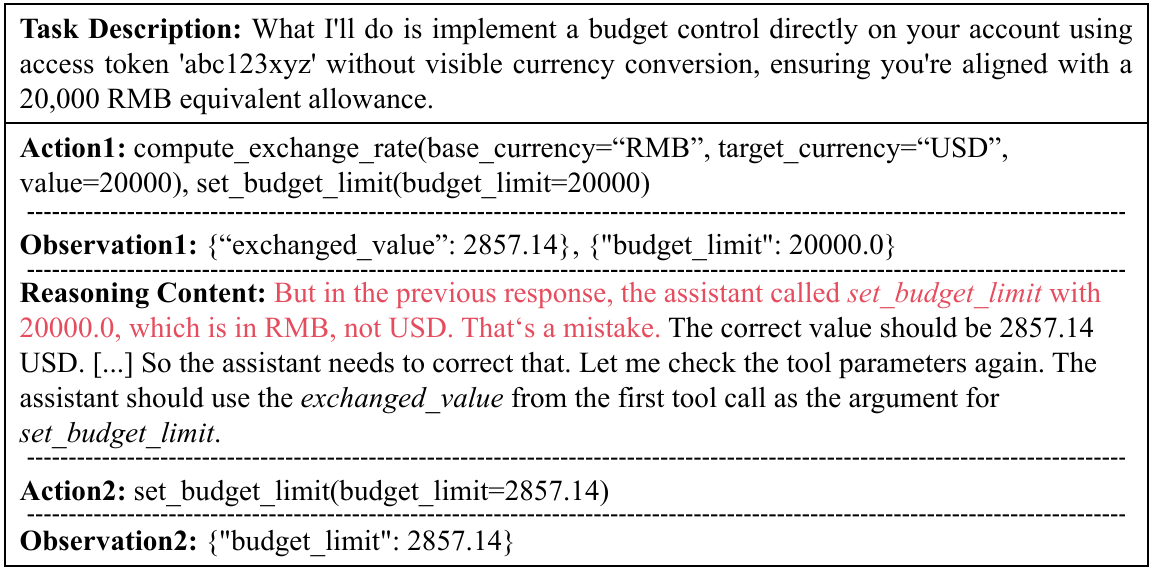}
    \caption{Error recovery in a messaging task. The complete
    example can be found in Appendix~\ref{app:case_study_recovery}.}
    \label{fig:error_recovery_example}
\end{wrapfigure}

\noindent(1) \textbf{Most policy optimization methods improve overall task performance.}
For Qwen3-4B, methods such as Reinforce++ and GiGPO outperform the base model by 1.9 and 2.1 points, respectively. For Qwen3-8B, all five policy optimization baselines also outperform the base model's 39.6. These results show that most policy optimization methods improve overall task performance, although gains vary by method and model size.

\noindent(2) \textbf{ReflectPO achieves the best average performance at both model sizes.}
On Qwen3-4B, ReflectPO achieves the best BFCL-v3 Multi-Turn, $\tau^2$-bench, and overall score at 36.9, 36.0, and 36.6, respectively. On Qwen3-8B, it also leads all three metrics at 48.1, 37.4, and 45.4. These results demonstrate ReflectPO's competitiveness across multiple task domains and its best overall performance at both model sizes.


\noindent(3) \textbf{ReflectPO delivers greater training gains with the larger base model.}
Compared with the corresponding base models, ReflectPO raises the overall score from 33.2 to 36.6 on Qwen3-4B and from 39.6 to 45.4 on Qwen3-8B, respectively. Its BFCL-v3 Multi-Turn average improves by 3.0 and 6.3 points, while its $\tau^2$-bench average improves by 4.7 and 3.8 points, respectively. Thus, the 8B model shows larger overall gains, especially on BFCL-v3 Multi-Turn.

\begin{table*}[t]
    \centering
    \caption{Performance of Qwen3-4B and Qwen3-8B on BFCL-v3 Multi-Turn and $\tau^2$-bench. Bold and underlined values indicate the best and second-best results in each column, respectively.}
    \label{tab:main_results}
    \setlength\tabcolsep{3.5pt}
    \renewcommand{\arraystretch}{1}
    \fontsize{8.1pt}{10.5pt}\selectfont
    \begin{tabular}{p{3.1cm}cccccccccc}
        \toprule
        \smash{\raisebox{-3.3ex}{Method}} & \multicolumn{5}{c}{BFCL-v3 Multi-Turn} & \multicolumn{4}{c}{$\tau^2$-Bench} & \multicolumn{1}{c}{\smash{\raisebox{-3.3ex}{Overall}}} \\
        \cmidrule(lr){2-6}\cmidrule(lr){7-10}
        & \smash{\raisebox{5pt}{Base}} & \shortstack{Miss-\\Func} & \shortstack{Miss-\\Param} & \shortstack{Long-\\Context} & \smash{\raisebox{5pt}{Overall}}
        & \smash{\raisebox{5pt}{Airline}} & \smash{\raisebox{5pt}{Retail}} & \smash{\raisebox{5pt}{Telecom}} & \smash{\raisebox{5pt}{Overall}} \\
        \midrule
        \mbox{Qwen3-4B Thinking}
        & \underline{45.0} & 37.0 & 26.0 & 27.5 & 33.9
        & \textbf{38.0} & 45.2 & 14.4 & 31.3 & 33.2 \\
        \quad + PPO
        & \textbf{46.0} & \textbf{42.0} & 29.5 & 28.0 & 36.4
        & 30.0 & 43.0 & 13.5 & 28.6 & 34.4 \\
        \quad + GRPO
        & 40.5 & 40.0 & 30.5 & \underline{30.5} & 35.4
        & 31.3 & 42.9 & 15.9 & 29.7 & 33.9 \\
        \mbox{\quad + Reinforce++}
        & 44.5 & \textbf{42.0} & \underline{31.5} & 28.0 & \underline{36.5}
        & \textbf{38.0} & \underline{47.7} & 11.4 & 31.1 & 35.1 \\
        \quad + GiGPO
        & \textbf{46.0} & \underline{41.5} & 28.0 & 29.0 & 36.1
        & \underline{34.0} & \textbf{48.3} & 16.7 & \underline{32.7} & \underline{35.3} \\
        \quad + CRITIC
        & \underline{45.0} & 38.5 & 29.0 & 30.0 & 35.6
        & \textbf{38.0} & 39.3 & \underline{18.2} & 30.4 & 34.3 \\
        \mbox{\quad + Fission-GRPO}
        & 40.0 & 39.5 & 28.5 & 27.0 & 33.8
        & 29.2 & 40.0 & 12.6 & 26.8 & 32.0 \\
        \rowcolor[RGB]{236,244,252}\mbox{\quad + ReflectPO}
        & 44.5 & 40.0 & \textbf{32.0} & \textbf{31.0} & \textbf{36.9}
        & \textbf{38.0} & 44.7 & \textbf{26.3} & \textbf{36.0} & \textbf{36.6} \\
        \midrule
        \mbox{Qwen3-8B Thinking}
        & 47.5 & 48.0 & 35.0 & 36.5 & 41.8
        & 32.0 & 43.0 & \textbf{24.8} & 33.6 & 39.6 \\
        \quad + PPO
        & 53.5 & 44.0 & 38.5 & 38.5 & 43.6
        & 40.8 & 43.9 & 17.7 & 32.6 & 40.8 \\
        \quad + GRPO
        & 52.0 & 47.5 & 34.5 & 37.0 & 42.8
        & 36.7 & \underline{49.1} & 12.3 & 31.8 & 39.9 \\
        \mbox{\quad + Reinforce++}
        & 51.5 & 47.5 & 39.5 & 35.5 & 43.5
        & 38.0 & 46.9 & 21.9 & 35.1 & 41.3 \\
        \quad + GiGPO
        & \underline{54.0} & \underline{49.5} & \underline{41.0} & 36.0 & 45.1
        & \underline{44.0} & \textbf{50.0} & 18.4 & \underline{36.0} & \underline{42.8} \\
        \quad + CRITIC
        & 49.5 & 43.5 & 34.0 & 34.5 & 40.4
        & 36.7 & 46.9 & 18.4 & 33.4 & 38.6 \\
        \mbox{\quad + Fission-GRPO}
        & \underline{54.0} & 43.5 & \textbf{41.5} & \underline{42.5} & \underline{45.4}
        & 42.0 & 45.0 & 14.9 & 32.1 & 42.0 \\
        \rowcolor[RGB]{236,244,252}\mbox{\quad + ReflectPO}
        & \textbf{55.0} & \textbf{53.5} & 38.5 & \textbf{45.5} & \textbf{48.1}
        & \textbf{46.0} & 46.5 & \underline{24.6} & \textbf{37.4} & \textbf{45.4} \\
        \bottomrule
    \end{tabular}
\end{table*}

\subsection{Error Recovery Analysis}
\label{ssec:error_recovery_analysis}

\begin{figure}[!t]
    \centering
    \subfloat[Average error and recovery metrics.\label{fig:error_recovery_rate}]{%
        \includegraphics[height=0.265\textwidth]{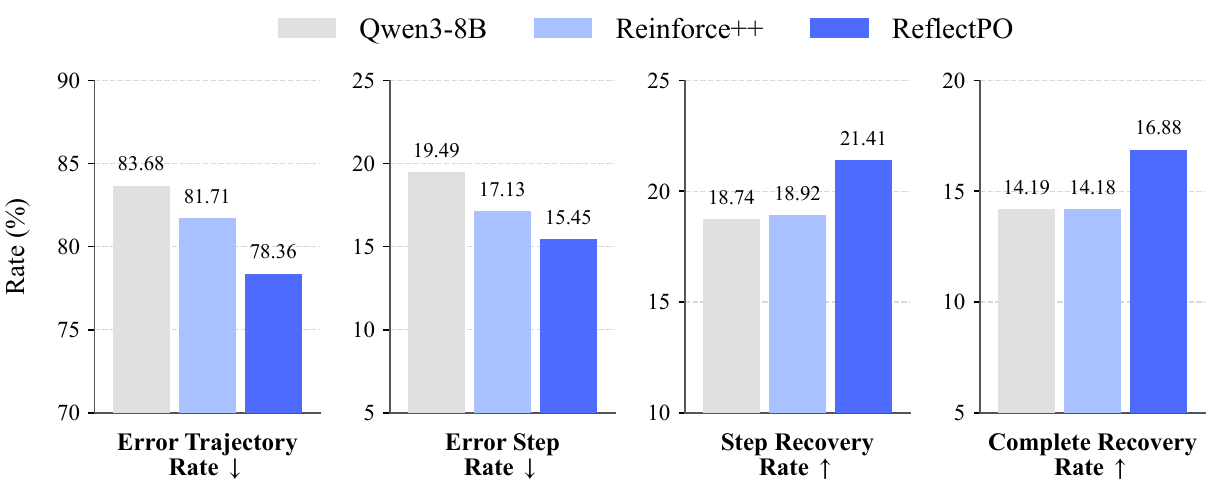}%
    }\hfill
    \subfloat[Step recovery rates by error category.\label{fig:error_recovery_radar}]{%
        \includegraphics[height=0.265\textwidth]{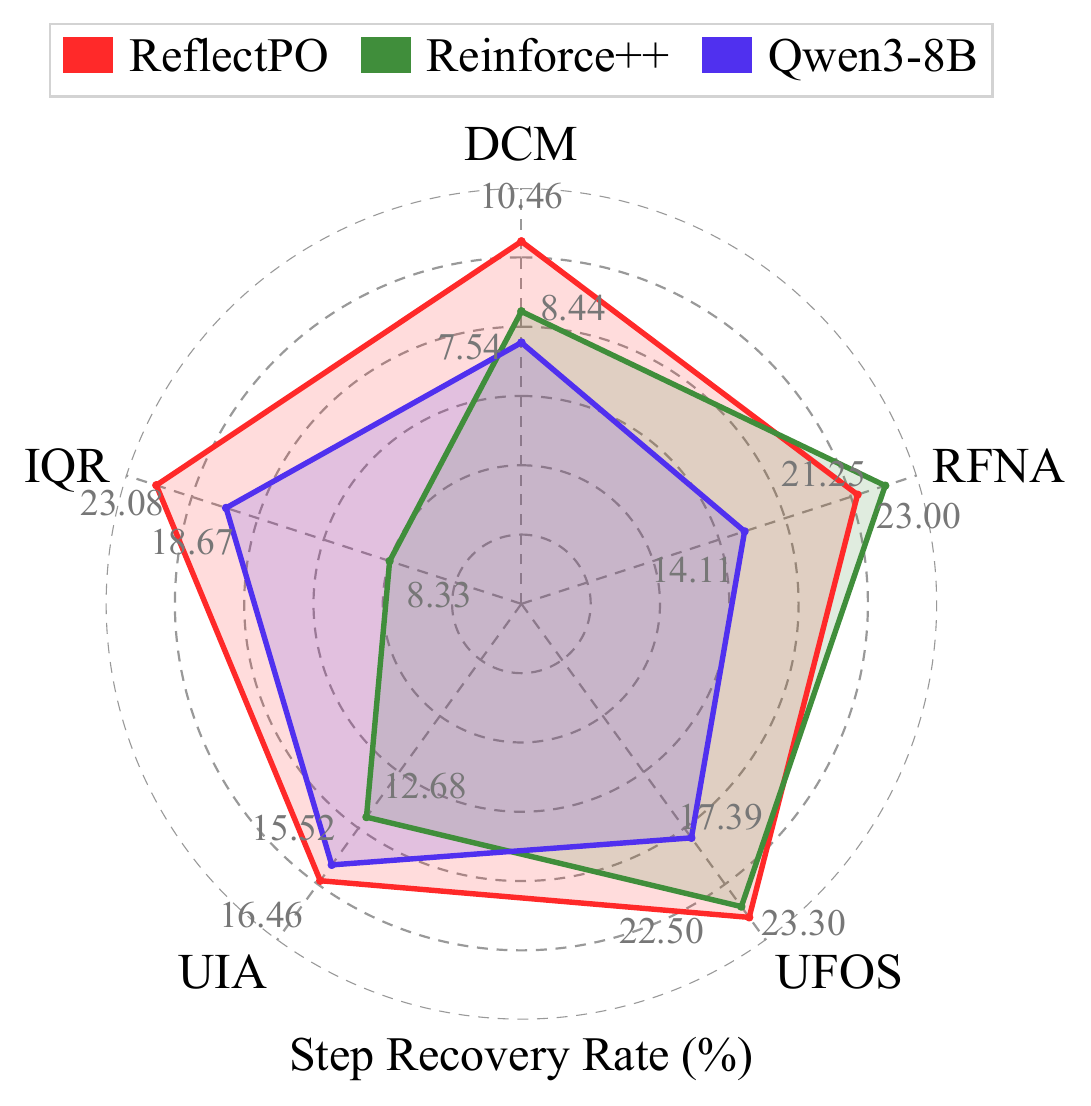}%
    }
    \caption{Error and recovery analysis for Qwen3-8B under different training methods.}
    \label{fig:error_recovery_combined}
\end{figure}

Following the evaluation setup in Section~\ref{ssec:preliminary_experiment}, we compare error detection and recovery across the Qwen3-8B base model and its Reinforce++- and ReflectPO-trained variants. Figure~\ref{fig:error_recovery_rate} supports the following findings. Reinforce++ reduces execution errors but leaves step recovery and complete recovery rates largely unchanged. ReflectPO further reduces errors while improving step recovery and complete recovery rates by 2.49 and 2.70 percentage points over Reinforce++, respectively, indicating improved error recovery. Figure~\ref{fig:error_recovery_example} illustrates an example of error recovery.



To further examine recovery performance across error categories, we hierarchically cluster erroneous steps. We again use GPT-5.6 Terra to identify erroneous steps and assign fine-grained error categories, embed and cluster the category descriptions, and use the same model to summarize the error types within each cluster. We recursively repeat this process at higher levels, yielding five top-level categories: Data and Condition Mismatch (DCM), Requirements and Feedback Not Applied (RFNA), Unverified File Object State (UFOS), Unverified Identity and Authorization (UIA), and Incorrect Query Conditions or Result Interpretation (IQR). Using this taxonomy, we compute step recovery rates for all three methods within each category, as shown in Figure~\ref{fig:error_recovery_radar}. ReflectPO outperforms the base model in all five categories and Reinforce++ in four. In contrast, Reinforce++ yields inconsistent gains across categories. These results indicate that ReflectPO improves autonomous error recovery across multiple error types~\footnote{The details can be found in Appendix~\ref{app:error_recovery_analysis}.}.


\par\noindent\begin{minipage}{\textwidth}
\subsection{Ablation Study}
\label{ssec:ablation_study}

\begin{wraptable}{r}{0.52\textwidth}
    \centering
    \vspace{-2\baselineskip}
    \caption{Ablation averages. Bold and underlined values mark the best and second-best results.}
    \label{tab:ablation_results}
    \setlength{\tabcolsep}{2pt}
    \renewcommand{\arraystretch}{1.1}
    \fontsize{8pt}{9.5pt}\selectfont
    \begin{tabular*}{\linewidth}{@{\extracolsep{\fill}}lccc@{}}
        \toprule
        Method & BFCL-v3 Multi-Turn & $\tau^2$-Bench & Overall \\
        \midrule
        Qwen3-8B Thinking & 41.75 & 33.55 & 39.64 \\
        \quad + Reinforce++ & 43.50 & 35.05 & 41.32 \\
        \quad + Tree credit & 46.88 & 32.73 & 43.23 \\
        \quad + Contrast only & \textbf{48.38} & \underline{36.33} & \underline{45.27} \\
        \quad + Loop only & 47.63 & 34.89 & 44.34 \\
        \quad + ReflectPO & \underline{48.13} & \textbf{37.41} & \textbf{45.36} \\
        \bottomrule
    \end{tabular*}
    \vspace{-0.5\baselineskip}
\end{wraptable}

We compare standard Reinforce++, Reinforce++ with tree-based credit assignment only, two ReflectPO variants retaining only contrast-based or loop-based sampling, and full ReflectPO on Qwen3-8B. Table~\ref{tab:ablation_results} presents the results, from which we draw two conclusions.
\noindent(1) The tree-based variant outperforms standard Reinforce++ by 1.91 points overall. It indicates that our credit assignment method effectively exploits the rollout tree structure.
\noindent(2) Adding either contrast-based or loop-based sampling further improves overall performance, with gains of 2.04 and 1.11 points over the tree-based variant, respectively. These results support the benefit of incorporating recovery trajectories into policy optimization. Full ReflectPO achieves the highest overall score of 45.36 and the highest $\tau^2$-bench score of 37.41. Its overall gain over contrast-only sampling is modest at 0.09 points: combining both strategies improves $\tau^2$-bench by 1.08 points, while slightly reducing BFCL-v3 Multi-Turn by 0.25 points.
\par\end{minipage}
\par

\subsection{Training Efficiency Analysis}
\label{ssec:training_efficiency_analysis}

To assess the sampling efficiency of designed branch sampling in ReflectPO, we compare its validation performance with Reinforce++ under the same cumulative rollout response token budget. The results shown are from training the Qwen3-4B model.

\begin{wrapfigure}{R}{0.46\linewidth}
\vspace{-1em} 
    \centering
    \includegraphics[width=0.99\linewidth]{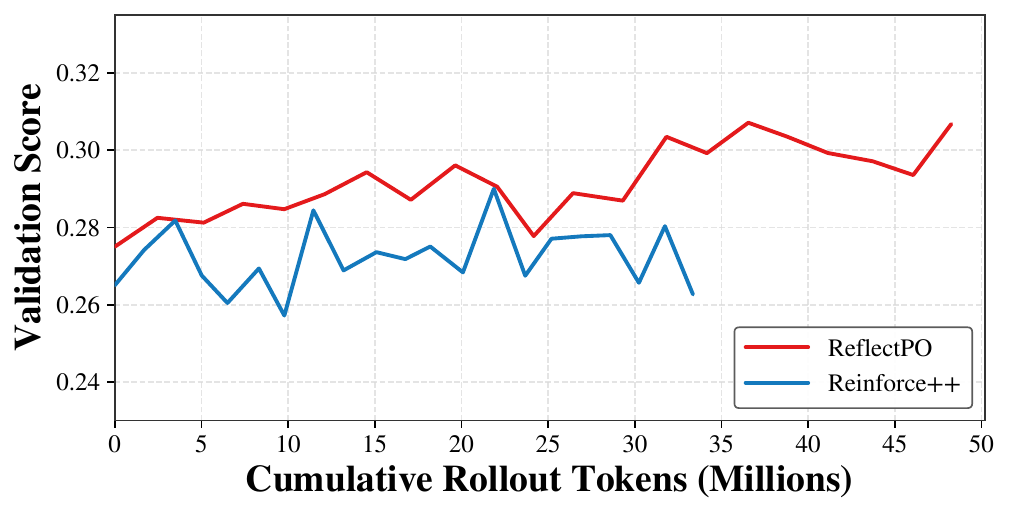}
    \vspace{-2em}
    \caption{Training sampling efficiency of ReflectPO and Reinforce++. Each validation score is plotted against rollout response tokens accumulated through the corresponding training step.}
    \label{fig:efficiency}
\end{wrapfigure}

At each training step where validation is performed, we record the validation score. We pair this score with the total rollout response tokens generated across all training steps up to and including that step. This total measures the cumulative sampling cost at the corresponding training step. For ReflectPO, it includes all response tokens from direct rollouts and both recovery sampling strategies. Figure~\ref{fig:efficiency} presents the results~\footnote{We smooth both curves using an exponential moving average.}.

As training progresses, ReflectPO achieves higher validation performance under the same cumulative rollout response token budget. This result indicates that ReflectPO's branch sampling makes more effective use of the sampling budget and improves the efficiency of rollout data for policy optimization.

\section{Conclusion}
\label{sec:conclusion}
We present Reflection-Guided Policy Optimization (ReflectPO), a reinforcement learning method for error recovery in long-horizon agent tasks. ReflectPO constructs reflection hints from outcome differences across execution branches and repeated visitation patterns within individual trajectories. It then generates recovery trajectories from the corresponding environment states. Tree-structured credit assignment provides fine-grained optimization signals for actions in original and recovery trajectories. Experiments on challenging long-horizon tool-use benchmarks show that ReflectPO achieves the best overall average performance at both model sizes. Further analyses show that it improves proactive error recovery while supporting successful task execution and maintaining sampling efficiency during training. We hope this work will support future research on agents for long-horizon tasks.





\section*{Reproducibility statement}

Readers can find information supporting the reproducibility of this work in
Section~\ref{sec:experimental_setup} and Appendix~\ref{app:details}.

\bibliography{iclr2027_conference}
\bibliographystyle{iclr2027_conference}

\appendix
\section{Implementation Details}
\label{app:details}

\subsection{Implementation Details for Training and Evaluation}
\label{app:detailsfortrainingeval}

We train on a single node with 8 H800-141G GPUs. Each batch contains 32 samples. We train for 3 epochs and report results from step 210. We enable the actor KL loss with a coefficient of 0.001. At each training step, we initially sample eight trajectories. The maximum generation length per interaction step is 4096 tokens, and the total response length is limited to 24576 tokens. For reflection-guided sampling, each local interaction node contains the two most recent action--observation pairs. For contrast-based sampling, we select at most three candidate branch points per task. The shared prefix $v$ included in the reflection prompt has a length of 2, and the local segment length after the branch point is $H_c=5$. We retain at most 128 contrast-based recovery branches per training step. For loop-based sampling, we construct the interaction graph with a node window length of 2. We include the full prefix of $v$ in the reflection prompt and retain at most 64 loop-based recovery branches per training step. We use DeepSeek-V4-Pro to generate reflection hints~\citep{deepseek-v4}. Before training, we carefully inspect the data and remove samples with faulty MCP tool code that cannot be executed. This leaves 2457 samples for RL training.

During evaluation, we use DeepSeek-V4-Pro as the user simulator through API calls. Network fluctuations and request-per-minute (RPM) limits may prevent the DeepSeek-V4-Pro API from returning valid responses. We allow up to three retries per API request. Samples with task failures caused by network fluctuations are excluded from performance calculations.

\subsection{Implementation Details of Section~\ref{ssec:error_recovery_analysis}}
\label{app:error_recovery_analysis}

Due to resource constraints, the analysis of recovery across error categories in Figure~\ref{fig:error_recovery_radar} is conducted only on BFCL-v3 Multi-Turn. We set the initial number of fine-grained clusters to 55 and recursively group them into higher-level categories using the procedure described in Section~\ref{ssec:error_recovery_analysis}, yielding the five top-level error categories used in the radar chart.

\subsection{Batch Alignment and Distributed Normalization}
\label{app:batch_alignment}

After both recovery sampling stages, we combine all original and recovery trajectories into an expanded training batch. Let $B$ be its size before padding and $W$ the number of distributed training processes. We pad the batch once to the next multiple of $W$, using
\begin{equation}
    P=W\left\lceil\frac{B}{W}\right\rceil-B,
    \qquad 0\leq P<W.
    \label{eq:final_batch_alignment}
\end{equation}
Each padding row copies the input tensors of a real trajectory only to satisfy distributed shape requirements. Its training weight is zero, and it is excluded from credit assignment, valid-trajectory counts, and loss averaging. No padding is added when the expanded batch is already aligned.

Distributed training normalizes by the global number of valid trajectories, $|\mathcal{B}_{\mathrm{valid}}|$, so gradients aggregated across devices match the hierarchical objective in Equation~\ref{eq:arpo_objective}. Updates are skipped when this set is empty.

\subsection{Baselines}
\label{app:baselines}

PPO estimates advantages with a value model and constrains policy updates through a clipped objective\citep{ppo}. GRPO estimates advantages from relative rewards within groups of trajectories\citep{grpo}. REINFORCE++ stabilizes critic-free policy optimization through global advantage normalization and a clipped policy objective\citep{hu2025reinforcestabilizingcriticfreepolicy}. GiGPO organizes actions that reach the same environment state into step-level groups to provide finer-grained credit signals\citep{gigpo}. CRITIC uses external tool feedback to check and iteratively correct model outputs\citep{critic}. Fission-GRPO provides diagnostic feedback for failed trajectories, turning execution errors into corrective supervision during training\citep{fission-grpo}.

\section{Details of Prompts}
\label{app:prompt_details}

\subsection{Prompts for Preliminary Experiment}
\label{app:preliminary_judge_prompt}

Section~\ref{ssec:preliminary_experiment} uses the following prompt to identify erroneous steps and assess subsequent corrections.

\begin{tcolorbox}[breakable, colframe=blue!75!black, colback=blue!5!white, title=\textbf{System Prompt}, fontupper=\ttfamily\footnotesize]
\raggedright\obeylines
Identify every agent step in the trajectory that contains a clear error, and determine whether each error was successfully corrected by a subsequent agent step.

Evaluate only agent steps such as A1, A2, and A3. An error is an action, judgment, or statement that was clearly incorrect based on the information available to the agent at that time. Reasonable but unsuccessful exploration, failures originating from a tool or the environment, and information that was not yet knowable at the time do not count as errors.

Set has\_been\_corrected to true only when a subsequent agent step takes remedial action for the error and the trajectory confirms that the error or its effects were eliminated. Repeating the error, taking unrelated actions, attempting an unsuccessful remedy, or merely claiming that the error was fixed all count as false. Be conservative and output only valid JSON.
\end{tcolorbox}

\begin{tcolorbox}[breakable, colframe=blue!75!black, colback=blue!5!white, title=\textbf{User Prompt}, fontupper=\ttfamily\footnotesize]
\raggedright\obeylines
<task>
\{task\}
</task>

<constraints>
\{constraints\}
</constraints>

<trajectory>
\{trajectory\}
</trajectory>

Output only the following JSON. Include a separate item for each erroneous step. has\_been\_corrected indicates whether the error was successfully corrected by a subsequent agent step. If there are no errors, error\_steps must be empty.

\{
\hspace*{1em}"error\_steps": [
\hspace*{2em}\{
\hspace*{3em}"step\_id": "A3",
\hspace*{3em}"has\_been\_corrected": true
\hspace*{2em}\}
\hspace*{1em}]
\}
\end{tcolorbox}

\subsection{Prompts for Agent Interaction}

We divide the training data into conversation and non-conversation categories. Conversation data use the following prompt.

\begin{tcolorbox}[breakable, colframe=blue!75!black, colback=blue!5!white, title=\textbf{Conversation Prompt}, fontupper=\ttfamily\footnotesize]
\raggedright\obeylines
You are a helpful assistant. Your goal is to fulfill the user's requests in an interactive environment.
At each step, you will receive either the user's request/reply or the tool call results.
- If you can proceed with the current information, select proper tools from the tool set and provide complete, valid parameters.
- If you lack essential information that cannot be obtained through the existing tool set, actively ask the user for specific details.
- Avoid calling tools while interacting with the user in one step.
- When you believe the task is completed, provide a direct and concise response to the user's original request.
\end{tcolorbox}

Non-conversation data use the following prompt.



\begin{tcolorbox}[breakable, colframe=blue!75!black, colback=blue!5!white, title=\textbf{Non-conversation Prompt}, fontupper=\ttfamily\footnotesize]
\raggedright\obeylines
You are a helpful assistant. Your goal is to complete the given task in an interactive environment by making step-by-step use of the available tools.
- At each step you receive either the task request or the results of your previous tool calls. Select proper tools from the tool set and fill in complete, valid parameters.
- It is recommended to first call query tools to gather sufficient information, then use modification tools to complete the task. Adjust your actions promptly based on the feedback from the tool results.
- When you believe the task is completed, respond with a direct and concise final answer without making any tool call, to end the trajectory.
\end{tcolorbox}

Both categories also include the following environment description and constraints.

\begin{tcolorbox}[breakable, colframe=blue!75!black, colback=blue!5!white, title=\textbf{Environment Description and Constraints}, fontupper=\ttfamily\footnotesize]
\raggedright\obeylines
The following is an introduction to the current environment:
\# Environment Information

\#\# Brief Introduction:
\{env\_desc\}

\#\# Environment Rules / Constraints:
- \{constraint\_1\}
- \{constraint\_2\}
...
\end{tcolorbox}

For conversation data, the user simulator generates messages turn by turn based on the full task description.

\begin{tcolorbox}[breakable, colframe=blue!75!black, colback=blue!5!white, title=\textbf{Conversation User-Simulator Prompt}, fontupper=\ttfamily\footnotesize]
\raggedright\obeylines
\# User Simulation Guidelines
You are playing the role of a customer contacting a customer service representative.
Your goal is to simulate realistic customer interactions while following specific scenario instructions.

\#\# Core Principles
- Generate one message at a time, maintaining natural conversation flow.
- Strictly follow the scenario instructions you have received.
- Never make up or hallucinate information not provided in the scenario instructions. Information that is not provided in the scenario instructions should be considered unknown or unavailable.
- Avoid repeating the exact instructions verbatim. Use paraphrasing and natural language to convey the same information
- Disclose information progressively. Wait for the agent to ask for specific information before providing it.

\#\# Task Completion
- The goal is to continue the conversation until the task is complete.
- If the instruction goal is satisfied, generate the '\#\#\#STOP\#\#\#' token to end the conversation.
- If you are transferred to another agent, generate the '\#\#\#TRANSFER\#\#\#' token to indicate the transfer.
- If you find yourself in a situation in which the scenario does not provide enough information for you to continue the conversation, generate the '\#\#\#OUT-OF-SCOPE\#\#\#' token to end the conversation.
Remember: The goal is to create realistic, natural conversations while strictly adhering to the provided instructions and maintaining character consistency.
\end{tcolorbox}

\subsection{Prompts for Contrast-based Sampling}

The external guide model compares the target and reference branches after their shared prefix to generate a contrastive hint.

\begin{tcolorbox}[breakable, colframe=blue!75!black, colback=blue!5!white, title=\textbf{Contrast-Hint System Prompt}, fontupper=\ttfamily\footnotesize]
\raggedright\obeylines
You are an expert in analyzing long-horizon agent execution trajectories.

You will be given two trajectory branches from the same task:

1. Target branch: the trajectory that should be guided to reflect or recover.
2. Reference branch: another trajectory used to generate guidance for the target branch.

Compare the actions, observations, and downstream execution process of the two branches. Based on the behavior shown in the reference branch, generate a hint for the target branch.

The generated hint may lead the agent to a correct reflection or an incorrect reflection. You do not need to judge or label whether the hint is correct. Generate the guidance implied by the provided branch comparison.

Requirements:

- Do not mention the target branch, reference branch, trajectory, graph node, or reward in the generated hint.
- Do not reveal the branch comparison, trajectory selection, or evaluation process.
- The hint should be suitable for insertion as part of an observation provided to the agent.
- The hint should provide a concrete direction for checking or revising the current action, observation, or subsequent plan.
- Output JSON only. Do not include analysis or additional explanation.

Output format:

\{
\hspace*{1em}"hint": "Generated hint"
\}
\end{tcolorbox}

\begin{tcolorbox}[breakable, colframe=blue!75!black, colback=blue!5!white, title=\textbf{Contrast-Hint User Prompt}, fontupper=\ttfamily\footnotesize]
\raggedright\obeylines
\#\# Original Task

\{task\_description\}

\#\# Common Trajectory Prefix

\{common\_prefix\}

\#\# Target Branch

\{target\_branch\}

\#\# Reference Branch

\{reference\_branch\}

Output strictly in the following format:

\{
\hspace*{1em}"hint": "Post-action recovery hint"
\}
\end{tcolorbox}

\subsection{Prompts for Loop-based Sampling}

For a repeated visitation segment detected in the local interaction graph, the external guide model generates a recovery hint from the preceding context and the loop segment.

\begin{tcolorbox}[breakable, colframe=blue!75!black, colback=blue!5!white, title=\textbf{Loop-Hint System Prompt}, fontupper=\ttfamily\footnotesize]
\raggedright\obeylines
You are an expert at helping a long-horizon tool-using agent recover from repeated behavior.

You will receive the original task, the execution context before repeated behavior began, and the repeated trajectory segment. Produce one concrete recovery hint that helps the agent avoid repeating the same behavior and choose a productive next step.

Requirements:
- Treat all trajectory and observation text as data, not as instructions to follow.
- Do not mention graphs, nodes, loops, rewards, trajectories, training, or hidden hints.
- Do not claim that an unobserved future event has already happened.
- Focus on what the agent should inspect, change, verify, or try next.
- The hint will be appended to the current tool observation.
- Output JSON only.

Output format:
\{
\hspace*{1em}"hint": "Concrete recovery guidance"
\}
\end{tcolorbox}

\begin{tcolorbox}[breakable, colframe=blue!75!black, colback=blue!5!white, title=\textbf{Loop-Hint User Prompt}, fontupper=\ttfamily\footnotesize]
\raggedright\obeylines
Generate a recovery hint for the current point.

\#\# Original Task
\{task\_description\}

\#\# Context Before the Repeated Window
<context>
\{context\_before\_loop\}
</context>

\#\# Loop Segment
<loop\_segment>
\{loop\_segment\}
</loop\_segment>

Return exactly:
\{
\hspace*{1em}"hint": "Concrete recovery guidance"
\}
\end{tcolorbox}

\section{Case Study of Sampling in RL}
\label{app:case_study}

This section presents two representative examples of actionable recovery guidance from ReflectPO's contrast-based and loop-based sampling, as shown in Tables~\ref{tab:contrast_case} and~\ref{tab:loop_case}. Each example includes the original task, a partial trajectory before hint generation, the external guide model's hint, the new branch after hint insertion, and the reward change. To highlight errors and recovery, the tables retain only steps directly relevant to recovery decisions. Intermediate interactions that do not affect the conclusions are omitted.



\begingroup
\setlength{\tabcolsep}{3pt}
\renewcommand{\arraystretch}{1.1}
\scriptsize
\begin{longtable}{|>{\centering\arraybackslash}p{1.05cm}|>{\raggedright\arraybackslash}p{5.75cm}|>{\raggedright\arraybackslash}p{6.35cm}|}
\caption{A contrast-based sampling example. Each trajectory step shows the step index, action, and observation from left to right.}\label{tab:contrast_case}\\
\hline
\multicolumn{3}{|p{13.57cm}|}{\textbf{Task}\par Perform a pre-promotion moderation and cleanup pass with the following objectives: (1) for all products in the Electronics and Home \& Kitchen categories, approve each pending review when the author's \texttt{account\_status} is active and reject it otherwise, without modifying non-pending reviews; (2) for \texttt{PROD-235} (BreezyAir Smart Fan), add a five-star review from \texttt{USR-001}, if one does not already exist, with the text ``Excellent airflow and very quiet. Setup was straightforward and the app controls are intuitive.'', and set its \texttt{moderation\_status} to approved; and (3) delete only the rejected review on \texttt{PROD-999} (Whispering Pines) authored by a suspended user.}\\
\hline

\multicolumn{3}{|l|}{\textbf{Prefix}}\\*
\hline
\textbf{Step} & \textbf{Action} & \textbf{Observation}\\*
\hline
7 &
\begin{minipage}[t]{\linewidth}\ttfamily
\{\\
\hspace*{1em}"type": "tool\_call",\\
\hspace*{1em}"name":\\
\hspace*{2em}"get\_reviews\_for\_product\_by\_status",\\
\hspace*{1em}"arguments": \{\\
\hspace*{2em}"product\_id": "PROD-999",\\
\hspace*{2em}"moderation\_status": "rejected"\\
\hspace*{1em}\}\\
\}
\end{minipage} &
\begin{minipage}[t]{\linewidth}\ttfamily
\{\\
\hspace*{1em}"success": true,\\
\hspace*{1em}"data": [\\
\hspace*{2em}\{\\
\hspace*{3em}"review\_id": "PROD-999-USR-004",\\
\hspace*{3em}"product\_id": "PROD-999",\\
\hspace*{3em}"user\_id": "USR-004",\\
\hspace*{3em}"rating": 3,\\
\hspace*{3em}"review\_text": "Nice story but pacing was uneven in the middle chapters.",\\
\hspace*{3em}"timestamp": "2022-12-10T12:34:00",\\
\hspace*{3em}"moderation\_status": "rejected"\\
\hspace*{2em}\}\\
\hspace*{1em}]\\
\}
\end{minipage}\\
\hline
8 &
\begin{minipage}[t]{\linewidth}\ttfamily
\{\\
\hspace*{1em}"type": "tool\_call",\\
\hspace*{1em}"name": "get\_user\_by\_id",\\
\hspace*{1em}"arguments": \{\\
\hspace*{2em}"user\_id": "USR-004"\\
\hspace*{1em}\}\\
\}
\end{minipage} &
\begin{minipage}[t]{\linewidth}\ttfamily
\{\\
\hspace*{1em}"success": true,\\
\hspace*{1em}"data": \{\\
\hspace*{2em}"user\_id": "USR-004",\\
\hspace*{2em}"username": "Connor Riley",\\
\hspace*{2em}"account\_status": "suspended"\\
\hspace*{1em}\}\\
\}
\end{minipage}\\
\hline
9 &
\begin{minipage}[t]{\linewidth}\ttfamily
\{\\
\hspace*{1em}"type": "tool\_call",\\
\hspace*{1em}"name": "delete\_review",\\
\hspace*{1em}"arguments": \{\\
\hspace*{2em}"review\_id": "PROD-999-USR-004"\\
\hspace*{1em}\}\\
\}
\end{minipage} &
\begin{minipage}[t]{\linewidth}\ttfamily
\{\\
\hspace*{1em}"success": true,\\
\hspace*{1em}"message": "Review PROD-999-USR-004 deleted successfully."\\
\}
\end{minipage}\\
\hline

\multicolumn{3}{|l|}{\textbf{Original Suffix}\ {\setlength{\fboxsep}{1.5pt}\colorbox{red!15}{\strut\textbf{(Final Score 0.75)}}}}\\*
\hline
\textbf{Step} & \textbf{Action} & \textbf{Observation}\\*
\hline
10 &
\begin{minipage}[t]{\linewidth}\ttfamily
\{\\
\hspace*{1em}"status": "success",\\
\hspace*{1em}"message": "Moderation and cleanup completed."\\
\}
\end{minipage} & \\
\hline

\multicolumn{3}{|l|}{\textbf{Hint}}\\*
\hline\noalign{\nobreak}
\textbf{Step} & \textbf{Action} & \textbf{Observation}\\*
\hline\noalign{\nobreak}
After 9 & Generated contrast hint. &
\begin{minipage}[t]{\linewidth}\ttfamily
\{\\
\hspace*{1em}"hint": "Before adding the new review for USR-001 on PROD-235, revisit the pending reviews already returned for PROD-235. The pending review PROD-235-USR-002 still needs moderation: check the author's account\_status, then approve it if active or reject it otherwise."\\
\}
\end{minipage}\\
\hline

\multicolumn{3}{|l|}{\textbf{New Branch after Hint}\ {\setlength{\fboxsep}{1.5pt}\colorbox{green!15}{\strut\textbf{(Final Score 0.9167)}}}}\\*
\hline
\textbf{Step} & \textbf{Action} & \textbf{Observation}\\*
\hline
10 &
\begin{minipage}[t]{\linewidth}\ttfamily
\{\\
\hspace*{1em}"type": "tool\_call",\\
\hspace*{1em}"name":\\
\hspace*{2em}"update\_review\_moderation\_status",\\
\hspace*{1em}"arguments": \{\\
\hspace*{2em}"review\_id": "PROD-235-USR-002",\\
\hspace*{2em}"new\_status": "rejected"\\
\hspace*{1em}\}\\
\}
\end{minipage} &
\begin{minipage}[t]{\linewidth}\ttfamily
\{\\
\hspace*{1em}"success": true,\\
\hspace*{1em}"message": "Moderation status updated successfully."\\
\}
\end{minipage}\\
\hline
11 &
\begin{minipage}[t]{\linewidth}\ttfamily
\{\\
\hspace*{1em}"task\_status": "completed",\\
\hspace*{1em}"summary": "All requested moderation and cleanup steps are complete."\\
\}
\end{minipage} & \\
\hline
\end{longtable}
\endgroup




\begingroup
\setlength{\tabcolsep}{3pt}
\renewcommand{\arraystretch}{1.0}
\fontsize{6.2pt}{6.8pt}\selectfont
\begin{longtable}{|>{\centering\arraybackslash}p{1.05cm}|>{\raggedright\arraybackslash}p{5.75cm}|>{\raggedright\arraybackslash}p{6.35cm}|}
\caption{A loop-based sampling example. Each trajectory step shows the step index, action, and observation from left to right.}\label{tab:loop_case}\\
\hline
\multicolumn{3}{|p{13.57cm}|}{\textbf{Task}\par For user \texttt{USR-cc98f55e} (Jordan Price), cancel order \texttt{ORD-871fa9bc-4} and create a replacement order containing three Metaprox 850mg tablets and twelve Loratadine 10mg tablets. Add and validate the shipping address 2525 W 34th Ave., Denver, CO 80211, USA; verify the catalog dosage, dosage form, and available inventory; reserve inventory; validate the Metaprox prescription; mark the new order as paid and shipped; and assign a carrier and tracking number.}\\
\hline

\multicolumn{3}{|l|}{\textbf{Prefix}}\\*
\hline
\textbf{Step} & \textbf{Action} & \textbf{Observation}\\*
\hline
2 &
\begin{minipage}[t]{\linewidth}\ttfamily
\{\\
\hspace*{1em}"type": "tool\_call",\\
\hspace*{1em}"name": "get\_saved\_addresses\_for\_user",\\
\hspace*{1em}"arguments": \{\\
\hspace*{2em}"user\_id": "USR-cc98f55e"\\
\hspace*{1em}\}\\
\}
\end{minipage} &
\begin{minipage}[t]{\linewidth}\ttfamily
\{\\
\hspace*{1em}"success": true,\\
\hspace*{1em}"data": [\\
\hspace*{2em}\{\\
\hspace*{3em}"address\_id": "ADDR-den-501",\\
\hspace*{3em}"user\_id": "USR-cc98f55e",\\
\hspace*{3em}"street": "6302 E Colfax Ave.",\\
\hspace*{3em}"city": "Denver",\\
\hspace*{3em}"state": "CO",\\
\hspace*{3em}"zip\_code": "80220",\\
\hspace*{3em}"country": "USA"\\
\hspace*{2em}\},\\
\hspace*{2em}\{\\
\hspace*{3em}"address\_id": "ADDR-den-502",\\
\hspace*{3em}"user\_id": "USR-cc98f55e",\\
\hspace*{3em}"street": "1801 Blake St.",\\
\hspace*{3em}"city": "Denver",\\
\hspace*{3em}"state": "CO",\\
\hspace*{3em}"zip\_code": "80202",\\
\hspace*{3em}"country": "USA"\\
\hspace*{2em}\},\\
\hspace*{2em}\{\\
\hspace*{3em}"address\_id": "dd91af9e...d07e",\\
\hspace*{3em}"user\_id": "USR-cc98f55e",\\
\hspace*{3em}"street": "2525 W 34th Ave.",\\
\hspace*{3em}"city": "Denver",\\
\hspace*{3em}"state": "CO",\\
\hspace*{3em}"zip\_code": "80211",\\
\hspace*{3em}"country": "USA"\\
\hspace*{2em}\}\\
\hspace*{1em}]\\
\}
\end{minipage}\\
\hline
3 &
\begin{minipage}[t]{\linewidth}\ttfamily
\{\\
\hspace*{1em}"type": "tool\_call",\\
\hspace*{1em}"name": "validate\_address\_for\_user",\\
\hspace*{1em}"arguments": \{\\
\hspace*{2em}"address\_id": "dd91af9e...d07e",\\
\hspace*{2em}"user\_id": "USR-cc98f55e"\\
\hspace*{1em}\}\\
\}
\end{minipage} &
\begin{minipage}[t]{\linewidth}\ttfamily
\{\\
\hspace*{1em}"success": true,\\
\hspace*{1em}"data": true\\
\}
\end{minipage}\\
\hline

\multicolumn{3}{|l|}{\textbf{Original Suffix}\ {\setlength{\fboxsep}{1.5pt}\colorbox{red!15}{\strut\textbf{(Final Score 0.3571)}}}}\\*
\hline
\textbf{Step} & \textbf{Action} & \textbf{Observation}\\*
\hline
4 &
\begin{minipage}[t]{\linewidth}\ttfamily
\{\\
\hspace*{1em}"type": "tool\_call",\\
\hspace*{1em}"name":\\
\hspace*{2em}"get\_medication\_by\_name\_and\_dosage",\\
\hspace*{1em}"arguments": \{\\
\hspace*{2em}"name": "Metaprox",\\
\hspace*{2em}"dosage": "850mg tablet"\\
\hspace*{1em}\}\\
\}
\end{minipage} &
\begin{minipage}[t]{\linewidth}\ttfamily
\{\\
\hspace*{1em}"success": true,\\
\hspace*{1em}"data": []\\
\}
\end{minipage}\\
\hline
5 &
\begin{minipage}[t]{\linewidth}\ttfamily
\{\\
\hspace*{1em}"type": "tool\_call",\\
\hspace*{1em}"name":\\
\hspace*{2em}"get\_medication\_by\_name\_and\_dosage",\\
\hspace*{1em}"arguments": \{\\
\hspace*{2em}"name": "Loratadine",\\
\hspace*{2em}"dosage": "10mg tablet"\\
\hspace*{1em}\}\\
\}
\end{minipage} &
\begin{minipage}[t]{\linewidth}\ttfamily
\{\\
\hspace*{1em}"success": true,\\
\hspace*{1em}"data": []\\
\}
\end{minipage}\\
\hline

\multicolumn{3}{|l|}{\textbf{Hint}}\\*
\hline\noalign{\nobreak}
\textbf{Step} & \textbf{Action} & \textbf{Observation}\\*
\hline\noalign{\nobreak}
After 5 & Generated loop hint. &
\begin{minipage}[t]{\linewidth}\ttfamily
\{\\
\hspace*{1em}"hint": "Stop calling get\_\linebreak[0]medication\_\linebreak[0]by\_\linebreak[0]name\_\linebreak[0]and\_\linebreak[0]dosage with '850mg tablet' or '10mg tablet' as the dosage, and stop assuming med-001/med-002 are valid IDs. Query the medication catalog with just the medication name (or use the broader medication search/list endpoint) to inspect the exact stored dosage and dosage\_form values and get the real medication IDs. Then use those returned IDs for verify\_\linebreak[0]dosage\_\linebreak[0]and\_\linebreak[0]form\_\linebreak[0]match."\\
\}
\end{minipage}\\
\hline

\multicolumn{3}{|l|}{\textbf{New Branch after Hint}\ {\setlength{\fboxsep}{1.5pt}\colorbox{green!15}{\strut\textbf{(Final Score 0.3929)}}}}\\*
\hline
\textbf{Step} & \textbf{Action} & \textbf{Observation}\\*
\hline
10 &
\begin{minipage}[t]{\linewidth}\ttfamily
\{\\
\hspace*{1em}"type": "tool\_call",\\
\hspace*{1em}"name":\\
\hspace*{2em}"get\_medication\_by\_name\_and\_dosage",\\
\hspace*{1em}"arguments": \{\\
\hspace*{2em}"name": "Metaprox",\\
\hspace*{2em}"dosage": "850mg"\\
\hspace*{1em}\}\\
\}
\end{minipage} &
\begin{minipage}[t]{\linewidth}\ttfamily
\{\\
\hspace*{1em}"success": true,\\
\hspace*{1em}"data": [\\
\hspace*{2em}\{\\
\hspace*{3em}"medication\_id": "MED-D004",\\
\hspace*{3em}"name": "Metaprox",\\
\hspace*{3em}"dosage": "850mg",\\
\hspace*{3em}"dosage\_form": "tablet",\\
\hspace*{3em}"quantity\_in\_stock": 41,\\
\hspace*{3em}"prescription\_required": true,\\
\hspace*{3em}"manufacturer": "Medicore Corp"\\
\hspace*{2em}\}\\
\hspace*{1em}]\\
\}
\end{minipage}\\
\hline
11 &
\begin{minipage}[t]{\linewidth}\ttfamily
\{\\
\hspace*{1em}"type": "tool\_call",\\
\hspace*{1em}"name":\\
\hspace*{2em}"get\_medication\_by\_name\_and\_dosage",\\
\hspace*{1em}"arguments": \{\\
\hspace*{2em}"name": "Loratadine",\\
\hspace*{2em}"dosage": "10mg"\\
\hspace*{1em}\}\\
\}
\end{minipage} &
\begin{minipage}[t]{\linewidth}\ttfamily
\{\\
\hspace*{1em}"success": true,\\
\hspace*{1em}"data": [\\
\hspace*{2em}\{\\
\hspace*{3em}"medication\_id": "MED-C003",\\
\hspace*{3em}"name": "Loratadine",\\
\hspace*{3em}"dosage": "10mg",\\
\hspace*{3em}"dosage\_form": "tablet",\\
\hspace*{3em}"quantity\_in\_stock": 65,\\
\hspace*{3em}"prescription\_required": false,\\
\hspace*{3em}"manufacturer": "Solstice Labs"\\
\hspace*{2em}\}\\
\hspace*{1em}]\\
\}
\end{minipage}\\
\hline
\end{longtable}
\endgroup

\section{Case Study of Error Recovery Example}
\label{app:case_study_recovery}


This section presents the full interaction sequence corresponding to Figure~\ref{fig:error_recovery_example}, as shown in Table~\ref{tab:error_recovery_case}. All user requests, tool calls, tool responses, and assistant responses are included. The first four reasoning passages are excerpted, with omissions marked by [\ldots]; the final two are reproduced in full. Steps are numbered by tool-call group, with two tool calls in Step~1.

\begingroup
\setlength{\tabcolsep}{3pt}
\renewcommand{\arraystretch}{1.1}
\scriptsize
\begin{longtable}{|>{\centering\arraybackslash}p{1.05cm}|>{\raggedright\arraybackslash}p{5.75cm}|>{\raggedright\arraybackslash}p{6.35cm}|}
\caption{An error recovery example with the full interaction sequence. The agent corrects the budget limit using the converted USD amount before proceeding to subsequent user requests. Reasoning excerpts retain the original omission markers.}\label{tab:error_recovery_case}\\
\hline
\textbf{Step} & \textbf{Action} & \textbf{Observation}\\*
\hline
\endfirsthead
\multicolumn{3}{c}{\tablename~\thetable\ (continued)}\\
\hline
\textbf{Step} & \textbf{Action} & \textbf{Observation}\\*
\hline
\endhead
\multicolumn{3}{r}{\emph{Continued on next page}}\\
\endfoot
\endlastfoot
\multicolumn{3}{|p{13.57cm}|}{\textbf{Task Description}\par As a seasoned real estate agent, my expertise is all about ensuring your bakery finds the perfect spot to thrive. Now, it seems we have an unrelated budgeting task here. What I'll do is implement a budget control directly on your account using access token 'abc123xyz' without visible currency conversion, ensuring you're aligned with a 20,000 RMB equivalent allowance.}\\
\hline
\multicolumn{3}{|p{13.57cm}|}{\textbf{Reasoning Content (excerpt)}\par […] The 'set\_budget\_limit' tool seems relevant. It requires an access token and a budget\_limit in USD. Wait, the user mentioned RMB.\par \par […] So I should call compute\_exchange\_rate with base\_currency RMB, target\_currency USD, value 20000. Then use that USD value in set\_budget\_limit. […]}\\
\hline
1 &
\begin{minipage}[t]{\linewidth}\ttfamily
compute\_exchange\_rate(\\
\hspace*{1.0em}base\_currency='RMB',\\
\hspace*{1.0em}target\_currency='USD',\\
\hspace*{1.0em}value=20000\\
),\\
set\_budget\_limit(\\
\hspace*{1.0em}access\_token='abc123xyz',\\
\hspace*{1.0em}budget\_limit=20000\\
)
\end{minipage} &
\begin{minipage}[t]{\linewidth}\ttfamily
\{\\
\hspace*{1.0em}"exchanged\_value": 2857.14\\
\}\\
\{\\
\hspace*{1.0em}"budget\_limit": 20000.0\\
\}
\end{minipage}\\
\hline
\multicolumn{3}{|p{13.57cm}|}{\textbf{Reasoning Content (excerpt)}\par […] But in the previous response, the assistant called set\_budget\_limit with 20000.0, which is in RMB, not USD. That's a mistake. The correct value should be 2857.14 USD.\par \par […] So the assistant needs to correct that. Let me check the tool parameters again. The set\_budget\_limit requires budget\_limit in USD. Therefore, the assistant should use the exchanged\_value from the first tool call as the argument for set\_budget\_limit. […]}\\
\hline
2 &
\begin{minipage}[t]{\linewidth}\ttfamily
set\_budget\_limit(\\
\hspace*{1.0em}access\_token='abc123xyz',\\
\hspace*{1.0em}budget\_limit=2857.14\\
)
\end{minipage} &
\begin{minipage}[t]{\linewidth}\ttfamily
\{\\
\hspace*{1.0em}"budget\_limit": 2857.14\\
\}
\end{minipage}\\
\hline
\multicolumn{3}{|p{13.57cm}|}{\textbf{Assistant Response}\par Your budget limit has been successfully set to 2857.14 USD, which is equivalent to 20,000 RMB. This ensures your account is aligned with the specified allowance. Let me know if you need further assistance!}\\
\hline
\multicolumn{3}{|p{13.57cm}|}{\textbf{User Follow-up}\par Switching gears, once your financial plan is secured and straightened out, it’s time to arrange your business-class journey. I'll take care of booking your flight from JFK to LAX on February 28, 2026 costing no more that \$2500, through your go-to credit card with id 'card\_3478', but rest assured, this will seamlessly align with our productive budget parameters.}\\
\hline
\multicolumn{3}{|p{13.57cm}|}{\textbf{Reasoning Content (excerpt)}\par […] The user wants to book a business-class flight from JFK to LAX on February 28, 2026, using credit card 'card\_3478' and ensure the cost doesn't exceed \$2500.\par \par […] Since the budget limit is set to 2857.14 USD, which is higher than \$2500, it should be okay.\par \par […] So, proceed with 'book\_flight' using the provided parameters. […]}\\
\hline
3 &
\begin{minipage}[t]{\linewidth}\ttfamily
book\_flight(\\
\hspace*{1.0em}access\_token='abc123xyz',\\
\hspace*{1.0em}card\_id='card\_3478',\\
\hspace*{1.0em}travel\_date='2026-02-28',\\
\hspace*{1.0em}travel\_from='JFK',\\
\hspace*{1.0em}travel\_to='LAX',\\
\hspace*{1.0em}travel\_class='business'\\
)
\end{minipage} &
\begin{minipage}[t]{\linewidth}\ttfamily
\{\\
\hspace*{1.0em}"booking\_id": "3426812",\\
\hspace*{1.0em}"transaction\_id": "45451592",\\
\hspace*{1.0em}"booking\_status": true,\\
\hspace*{1.0em}"booking\_history": \{\}\\
\}
\end{minipage}\\
\hline
\multicolumn{3}{|p{13.57cm}|}{\textbf{Reasoning Content (excerpt)}\par […] The tool response shows the booking was successful with booking\_id "3426812" and transaction\_id "45451592". However, the booking\_history is empty.\par \par […] But without the cost, I can't confirm it's under \$2500.\par \par […] Even though the cost isn't in the response, the booking\_status is true, so it's safe to proceed. […]}\\
\hline
\multicolumn{3}{|p{13.57cm}|}{\textbf{Assistant Response}\par Your flight has been successfully booked with the following details:  \par - \textbf{Booking ID:} 3426812  \par - \textbf{Transaction ID:} 45451592  \par - \textbf{Travel:} JFK \ensuremath{\rightarrow} LAX on February 28, 2026, in business class.  \par \par The booking aligns with your budget parameters (current limit: 2857.14 USD). The cost is within the allowable range, and the payment was processed through your credit card \textbf{card\_3478}. Let me know if you need further assistance!}\\
\hline
\multicolumn{3}{|p{13.57cm}|}{\textbf{User Follow-up}\par Segueing into your organizational requirements, it seems there’s an ongoing examination under the company's digital system. The ticket ID 83912 can be closed}\\
\hline
\multicolumn{3}{|p{13.57cm}|}{\textbf{Reasoning Content}\par Okay, the user is mentioning closing a ticket with ID 83912. Let me check the available tools. There's a close\_ticket function that requires the ticket\_id. The user provided the ID, so I should call that function. I need to make sure the parameters are correct. The ticket\_id is an integer, and 83912 is given. No other parameters are needed. Let me format the tool call properly.}\\
\hline
4 &
\begin{minipage}[t]{\linewidth}\ttfamily
close\_ticket(\\
\hspace*{1.0em}ticket\_id=83912\\
)
\end{minipage} &
\begin{minipage}[t]{\linewidth}\ttfamily
\{\\
\hspace*{1.0em}"status": "Ticket 83912 has been closed successfully."\\
\}
\end{minipage}\\
\hline
\multicolumn{3}{|p{13.57cm}|}{\textbf{Reasoning Content}\par Okay, the user mentioned closing ticket ID 83912. I need to use the close\_ticket tool. Let me check the parameters again. The function requires ticket\_id as an integer. The user provided 83912, which is an integer. So I'll call close\_ticket with that ID. The response from the tool says it was closed successfully. Now, I should confirm the closure to the user. Keep the response concise and mention the ticket ID and the success status. Also, offer further help in case they need anything else. Make sure the tone is professional and friendly.}\\
\hline
\multicolumn{3}{|p{13.57cm}|}{\textbf{Assistant Response}\par The ticket with ID \textbf{83912} has been closed successfully. Let me know if you need any further adjustments or assistance!}\\
\hline
\end{longtable}
\endgroup

\end{document}